\documentclass[conference]{IEEEtran}
\IEEEoverridecommandlockouts
\usepackage{cite}
\usepackage{amsmath,amssymb,amsfonts}
\usepackage{algorithmic}
\usepackage{graphicx}
\usepackage{textcomp}
\usepackage{xcolor}
\usepackage{todonotes}
\usepackage{hyperref}
\usepackage{svg}
\usepackage{amsfonts}
\usepackage{orcidlink} 
\usepackage[vlined, ruled]{algorithm2e} 
\usepackage{orcidlink} 

\DeclareMathOperator*{\argmin}{argmin}
\DeclareMathOperator*{\argmax}{argmax}

\def\BibTeX{{\rm B\kern-.05em{\sc i\kern-.025em b}\kern-.08em
    T\kern-.1667em\lower.7ex\hbox{E}\kern-.125emX}}
\begin{document}

\title{Learning Causal Graphs in Manufacturing Domains using Structural Equation Models
}



\author{

\IEEEauthorblockN{Maximilian Kertel}
\IEEEauthorblockA{\textit{Technology Development Battery Cell} \\
\textit{BMW Group}\\
Munich, Germany \\
maximilian.kertel@bmw.de \orcidlink{0000-0003-3996-0642}}
\and
\IEEEauthorblockN{Stefan Harmeling}
\IEEEauthorblockA{\textit{Department of Computer Science} \\
\textit{TU Dortmund University}\\
Dortmund, Germany \\
stefan.harmeling@tu-dortmund.de
}
\and
\IEEEauthorblockN{Markus Pauly}
\IEEEauthorblockA{\textit{Department of Statistics} \\
\textit{TU Dortmund University}\\
Dortmund, Germany \\
\textit{Research Center Trustworthy}  \\
\textit{Data Science and Security} \\
UA Ruhr, Germany \\
pauly@statistik.tu-dortmund.de \orcidlink{0000-0002-0976-7190}}
}
\maketitle

\begin{abstract}
Many production processes are characterized by numerous and complex cause-and-effect relationships. Since they are only partially known they pose a challenge to effective process control. In this work we present how Structural Equation Models can be used for deriving cause-and-effect relationships from the combination of prior knowledge and process data in the manufacturing domain. Compared to existing applications, we do not assume linear relationships leading to more informative results.
\end{abstract}

\begin{IEEEkeywords}
Causal Discovery, Bayesian Networks, Industry 4.0
\end{IEEEkeywords}

\section{Introduction}
\textbf{To be published in the Proceedings of IEEE AI4I 2022.} \\
Complex manufacturing processes as, e.g. for battery cells show high scrap rates and thus high production costs and large environmental footprints. One of the driving factors is the missing knowledge on the interdependencies between the process parameters, intermediate product properties and the quality characteristics 
\cite{kornas2019data}. Together we call this the cause-and-effect relationships (CERs). CERs can be visualized as a network with the process and product characteristics as nodes and the CERs as directed edges \cite{kornas2019data, wuest2014approach}. It is the goal of our paper to unify expert knowledge and process data to derive such a network, which allows the visual identification of 
\begin{itemize}
\item root-causes of erroneous products,
\item relevant parameters for process control during successive production steps and
\item important characteristics to predict the quality of the final product.
\end{itemize}
In complex manufacturing domains, CERs form a linked mesh of hundreds of involved factors \cite{kornas2019data}. Typically, CERs are derived by running Designs of Experiments (DOEs). However, DOEs can be time-demanding and the production line has to be stopped in the meantime leading to prohibitively high costs. Moreover, if there are many potential CERs, the number of experiments can become infeasible. \\ 
At the same time, the Internet of Things (IoT) allows data processing and storage along the whole production line, leading to a vast amount of accessible information. It is thus desirable to derive the CERs from the existing observational (or non-experimental) data. For this purpose, Bayesian Networks can be used to unify expert knowledge and data. From these, CERs can be derived under the assumption of causal sufficiency \cite{SpirtesPC}. This approach is called \textit{causal discovery} or \textit{structure learning}. \\ 
The most common example in the manufacturing domain \cite{ReviewCausalDiscoveryManufacturing, marazopoulouCausal, LiStructureLearning}, is the PC algorithm \cite{SpirtesPC}. This algorithm relies on the assumption of faithfulness and on efficient statistical tests for conditional independence. 
In principle the PC algorithm can be applied with any test for conditional independence. However, existing nonparametric tests do not scale well \cite{ZhangKernelTest, fukumizu2007kernel}. 
Most of the applications of the PC algorithm either discretize the measurements, or researchers approximate the joint distribution of the variables by a multivariate normal distribution. For discrete data and normally distributed data fast tests for conditional independence exist. However, the former leads to a loss of information, while the latter requires a linear dependency between the variables to be exact. In case of manufacturing data this is most likely a misspecification \cite{ManufacturingIoT}. Simulation studies show, that the performance of the PC algorithm can be poor in case of non-linearity \cite{CAM}. This questions the application of the PC algorithm for large or high-dimensional manufacturing data. \\ 
In recent years, Structural Equation Models (SEM), which can incorporate arbitrary functional relationships, were increasingly proposed to derive Causal Bayesian Networks. They replace the assumption on faithfulness by a functional form of the conditional distributions (see Equation (\ref{eq:SEMGeneral})). While the PC algorithm returns a set of graphs, methods based on SEMs often derive a single graph. To the best of our knowledge, we are the first to apply SEMs to derive such graphical models in the manufacturing domain. \\
The paper is structured as follows. In Section \ref{sec:Manufacturing} we present potential prior knowledge and available data in manufacturing domains. We continue in Section \ref{sec:GraphicalModels} by reviewing Bayesian Networks and SEMs and explain Causal Additive Models (CAM). In Section \ref{sec:Methodology} we present an extension of CAM, called TCAM, which efficiently incorporates prior knowledge. 
We apply our method in Section \ref{sec:Application} to process data of the assembly of battery modules at BMW. We conclude in Section \ref{sec:Conclusion}.

\section{Data and Challenges in Complex Manufacturing Domains}\label{sec:Manufacturing}
In this section we describe the data sources and propose a preprocessing of the data. Then, we explain the broad prior knowledge in manufacturing domains. Finally, we mention common challenges with production data.
\subsection{Data Sources along the Production Line}
\begin{figure*}
\begin{center}
\includegraphics[width=\linewidth]{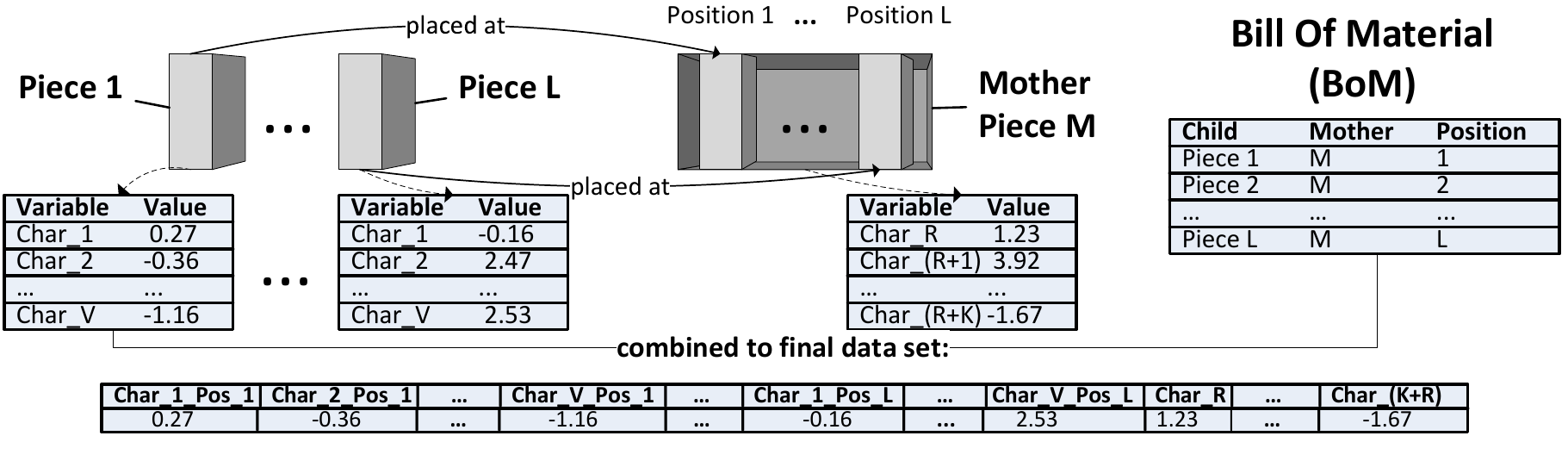}
\caption{Visualization of the data preparation described in Section \ref{sec:Manufacturing}. The same measurements are collected for piece $1$ to piece $L$. Then they are placed in their mother piece with identifier $M$. Finally, the resulting data set consists of all measurements of $M$ and those from piece $1$ to piece $L$, where the positioning of the measurements of the child pieces within the data frame depends on their placement according to the BoM. This step is carried out repeatedly, if $M$ itself is positioned in another mother piece.}
\label{fig:dataPrep}

\end{center}
\end{figure*}

The assembly of products consists of production lines, which again contain several stations, which are passed in a fixed order and where process steps are carried out. During those process steps the piece is transformed or it is combined with other parts in order to achieve a predefined outcome. All involved parts are assigned to unique identifiers. Data of different types is collected along the production process: 
\begin{itemize}
\item Process data: the stations take measurements of the involved parts (e.g. thickness of the piece) and the parameters of the machine (e.g. weight of applied glue). 
\item End-of-Line (EoL) tests take additional quality measurements of the intermediate or final products. 
\item Station information: at some production steps the pieces are spread out to identical stations, such that parts can be processed in parallel and every piece is assigned to one of the stations.
\item Bill of Material (BoM): the BoM contains the information which pieces were merged together and on which position they have been worked in.
\item Supplier data: suppliers transmit data on provided goods.
\end{itemize}
The preprocessing of the data, which is depicted in Figure \ref{fig:dataPrep}, consists of the following steps:
\begin{enumerate}
\item Collect the data for every intermediate product.
\item Iteratively merge the data of all subcomponents of a final product.
\end{enumerate}
Measurements of identical subcomponents, which are placed in the same position, can be found in the same column. Eventually, the final tabular data set contains all measurements that can be associated with a final product.
\subsection{Prior Knowledge}
As the stations are passed in a fixed order, we know that CERs across different stations can only act forward in time. \\
Additonally, in many manufacturing organizations, tools as the Failure Mode and Effect Analysis (FMEA) \cite{stamatis2003failure} are implemented to extract expert knowledge on CERs in the production process and to provide the information in a structured form. 
\subsection{Challenges of Data Analysis in Manufacturing}
Often, similar information is recorded multiple times along the production line, leading to multicollinearity \cite{ReviewCausalDiscoveryManufacturing}. Also, sensors might deliver non-informative data by recording implausible values. Industrial data is also reported to be drifting over time. However, even in shorter time intervals, data of a series production contains thousands of observations. This distinguishes the manufacturing domain from other applications of causal discovery as medicine, genetics or the social sciences. 



\section{Structure Learning of Graphical Models} \label{sec:GraphicalModels}
\subsection{Some Preliminaries on Graphical Models}
Let $G = \left(\mathbf{V}, \mathbf{E}\right)$ be a directed acyclic graph (DAG) \cite[Chapter~6]{PetersBook} with nodes $\mathbf{V} = \left(V_1, \ldots, V_p\right)$ and edges $\mathbf{E}$. The node $V_i$ is called a parent of $V_j$ if the edge $V_i \rightarrow V_j$ is in $\mathbf{E}$. We denote the set of all parents of $V_j$ as $pa(V_j)$. A tuple of nodes $(V_{j_1}, \ldots, V_{j_\ell})$, such that $V_{j_k}$ is a parent of $V_{j_{k+1}}$ for all $k=1, \ldots, (\ell-1)$, is called a \textit{directed path}. Nodes that can be reached from $X_j$ through a directed path are called the \textit{descendants} of $X_j$. \\
In the following we denote random vectors with bold letters as $\mathbf{Z}$ and random variables as $Z$. Let $\mathbf{X} = \left(X_1, \ldots, X_p\right)$ be a random vector representing the data generating process. For a graph $G$ with nodes $X_1, \ldots, X_p$, we call $\left(\mathbf{X}, G\right)$  a Bayesian network if the local Markov property holds, i.e.
\begin{equation*}
X_i \perp X_j | pa(X_i)
\end{equation*}
for any $X_j$ that is not a descendant of $X_i$ in $G$. Here, $X \perp Y | \mathbf{Z}$ denotes the conditional independence of $X$ and $Y$ given $\mathbf{Z}$. In that case, we can deduce additional conditional independencies for $\mathbf{X}$ from the graph $G$ using the concept of \textit{d-separation} \cite{PetersBook}. For a Bayesian Network $(\mathbf{X}, G)$, it then holds that $X_i \perp X_j | \mathbf{S}$ if $X_i$ and $X_j$ are d-separated by $\mathbf{S}$ in $G$. On the other hand, if there is a graph $G$, such that $X_i \perp X_j | \mathbf{S}$ implies that $X_i$ and $X_j$ are d-separated given $\mathbf{S}$ in $G$, then $\mathbf{X}$ is called \textit{faithful with respect to} $G$. As multiple graphs can contain the same d-separations, this graph $G$ is in general not unique. \\ 
To promote the intuition, assume that $\mathbf{X}$ has a joint density $f$. Then $X_i \perp X_j | \mathbf{S}$ can be characterized by
$$f\left(x_i | X_j = x_j, \mathbf{S} = \mathbf{s}\right) = f\left(x_i | \mathbf{S} = \mathbf{s}\right), $$
where $f\left(x_i | \mathbf{Z} = \mathbf{z}\right)$ denotes the conditional density function of $X_i$ given $\mathbf{Z} = \mathbf{z}$. Thus, if we already know $\mathbf{S}$, then $X_j$ does not provide additional information on $X_i$. 
Assume that we are interested which variable in $\{X_j, \mathbf{X}_\mathbf{S}\}$ causes the variable $X_i$ to be out of the specification limits. Then we know, that the root causes can be found within $\mathbf{S}$. 
\subsection{Graph Learning with Structural Equation Models}\label{sec:graphLearningSEM}
While the PC algorithm is the classic approach for deriving a Causal Bayesian Network, recent research focused on identifying it using acyclic SEMs  \cite{CAM, DAGsNoTears, Lingam, PetersAdditiveModelIdentifiable}. 
They assume that there exists a permutation $\Pi^0(1, \ldots, p) = \left( \pi^0(1), \ldots, \pi^0(p)\right)$ and functions $\{f_\ell, \ell = 1, \ldots, p\}$, such that
\begin{equation}\label{eq:SEMGeneral}
X_{\ell} = f_\ell(X_{\ell_1}, \ldots, X_{\ell_{v}}, \varepsilon_{\ell}), ~ \ell = 1, \ldots, p,
\end{equation}
where $\pi^0(\ell_k) < \pi^0(\ell)$ for all $k =1, \ldots, v$ and $\varepsilon_1, \ldots, \varepsilon_p$ are i.i.d. noise terms. As the estimation of $f_\ell$ in Equation (\ref{eq:SEMGeneral}) is difficult in high dimensions, one typically restricts the function class and the distribution of the noise terms. In this work, we assume that the functions follow the additive form
\begin{equation}\label{eq:CAM}
f_\ell(X_{\ell_1}, \ldots, X_{\ell_{v}}, \varepsilon_{\ell}) = c_\ell + \sum_{k: \pi^0(k) < \pi^0(\ell)} f_{k, \ell}(X_{k}) + \varepsilon_{\ell}, 
\end{equation}
where $\varepsilon_\ell \sim \mathcal{N}(0, \sigma_\ell)$ and $c_\ell \in \mathbb{R}$. To ensure the uniqueness of the $f_{k, \ell}$ and without loss of generality, we set $\mathbf{E}\left(X_{\ell}\right) = 0$ and $\mathbf{E}\left(f_{k, \ell}(X_{k})\right) = 0$, for all $\ell = 1, \ldots, p, \pi^0(k) < \pi^0(\ell)$. From Equations (\ref{eq:SEMGeneral}) and (\ref{eq:CAM}) we derive that 
$$X_\ell \perp X_k | \left(X_{v_1}, \ldots, X_{v_j}\right),$$ 
with $\pi^0(k) < \pi^0(\ell), \pi^0(v_1) < \pi^0(\ell), \ldots, \pi^0(v_j) < \pi^0(\ell)$ if and only if $f_{k, \ell} = 0$. Let $G^0$ be the graph on $X_1, \ldots, X_p$, that contains the edge $X_i \rightarrow X_j$ if and only if $f_{i, j} \neq 0$ for $\pi^0(i) < \pi^0(j)$. Then $(\mathbf{X}, G^0)$ is a Bayesian network, as it is fulfilling the Markov property. \\ 
If we assume that the functions $f_{k, \ell}$ in Equation (\ref{eq:CAM}) are non-linear and smooth, then \cite{PetersAdditiveModelIdentifiable} show that $G^0$ is identifiable from observational data. This is in contrast to the PC algorithm, which typically returns a class of graphs. Note that we do not presume that the distribution is faithful to some DAG, which is a central assumption of the PC algorithm. 
We emphasize that for the PC algorithm non-linearity is an obstacle as efficient conditional independence testing is just feasible for multivariate normal data. In contrast, we can utilize the non-linearity for identifying SEMs to receive more informative results (under the assumption of Equation (\ref{eq:CAM})). \\ 
An example of a learning algorithm for SEMs is the Causal Additive Model (CAM, \cite{CAM}). We will focus on CAM due to its applicability to high-dimensional data, its ability to capture non-linearity and due to the theoretical justification that $G^0$ can be identified, if the functions on the right-hand side of Equation \ref{eq:CAM} are nonlinear and smooth. \cite{CAM} propose to find $G^0$ with the following steps:  

\begin{enumerate}
\item Find the underlying node ordering $\Pi^0$ of $X_1, \ldots, X_p$.
\item Identify the influential functions $f_{k, \ell}$ with feature selection methods.
\end{enumerate}
To make things more precise, consider $N$ observations $\left(x_{i 1}, \ldots, x_{ip}\right), i = 1, \ldots, N$ from $\mathbf{X}$ and call the data matrix $\mathbf{D} \in \mathbb{R}^{N \times p}$.
\subsubsection{Finding the node ordering}
\cite{CAM} show that if 
\begin{itemize}
\item the functions $f_{k, \ell}$ are smooth and non-linear and can be approximated well and
\item the derivatives of $f_{k, \ell}$ and the fourth moments of $f_{k, \ell}(X_k)$ and $X_k$  are bounded.
\end{itemize}
then the following estimator for $\Pi^0$ is consistent as $N \rightarrow \infty$:

\begin{equation}\label{eq:OrderScore}
\hat{\Pi} = \argmin_\Pi \sum_{\ell = 1}^p ||x_{\ell} - \sum_{\pi(k) < \pi(\ell)} \widehat{f}_{k,\ell}(x_{k})||^2_{2, N}
\end{equation}
Here, we define $|| x_k ||_{2, N}^2 := \frac{1}{N}\sum_{k = 1}^N x_{k\ell}^2$ and $\widehat{f}_{k,\ell}$ is found by running an additive model regression \cite{WoodBookGAM} of $X_\ell$ on $\{X_{k}: ~ \pi(k) < \pi(\ell) \}$. \\
For large $p$, \cite{CAM} propose a greedy method to find $\hat{\Pi}$. Let $G$ be a DAG on $\mathbf{X}$ with edges $E(G)$. For simplicity we denote the edge $X_k \rightarrow X_\ell$ by $(k, \ell)$. A score for $G$ is defined by
$$S(G) = \sum_{\ell = 1}^p  ||x_{\ell} - \sum_{(k,l) \in E(G)} \widehat{f}_{k,\ell}(x_{k})||^2_{2, N}.$$
The functions $\widehat{f}_{k,\ell}$ are estimated by running an additive model regression of $X_\ell$ on its parents in $G$. Intuitively, $S(G)$ indicates how much variation of $\mathbf{D}$ is captured by $G$. The edges that can be added to $G$ without causing cycles are denoted by 
\begin{align*}
\begin{split}
A(G) := \{&(i,j) \in \{1, \ldots,p\} \times \{1, \ldots,p\}: \\ &(\mathbf{X}, E(G) \cup \{(i,j)\}) \text{ is DAG}\}.
\end{split}
\end{align*} 
Starting with the empty graph $G_0$, \cite{CAM} iteratively add the edge $({k^0}, {\ell^0}) = \argmax_{(k',\ell') \in A(G_t)} M_t(k', \ell')$, where
\begin{align}
\begin{split}\label{eq:ScoreImprove}
M_t(k', \ell') &= ||x_{\ell} - \sum_{(k,\ell) \in E(G_t)} \widehat{f}_{k,\ell}(x_{k})||^2_{2, N} \\ &- ||x_{\ell} - \sum_{(k,l) \in E(G_t) \cup \{(k', \ell')\}}  \widetilde{f}_{k,\ell}(x_{k})||^2_{2, N}.
\end{split}
\end{align}
The functions $\widehat{f}_{k,\ell}$ are found by regressing $X_\ell$ on its parents in $G_t$, while $\widetilde{f}_{k,\ell}$ are found by regressing $X_\ell$ on its parents in $G' = (\mathbf{X}, E(G_t) \cup \{(k',\ell')\})$. Thus, the edge $(k^0, \ell^0)$ maximally reduces the unexplained variance. We set $G_{t+1} = (\mathbf{X}, E(G_t) \cup \{(k^0,\ell^0)\})$ 
and continue until we obtain a complete DAG, which implies the node ordering. \\ 
This greedy method is still computationally intense for large $p$. Thus, \cite{CAM} propose to take advantage of sparse structures, where $p$ is large but the number of edges in the graph is assumed to be small: to this end they start by a preliminary neighborhood selection (PNS) step. Here, initially for every $\ell \in \{1, \ldots, p\}$ a superset of neighbors of $X_\ell$ in $G^0$ is identified. In the subsequent node ordering step, one only considers the superset of the neighbors, when greedily adding new edges. This reduces the computation time of the algorithm significantly, if the sizes of the supersets are considerably smaller than $p$. \\ 
\subsubsection{Identifying edges}
After the node ordering is set, we need to identify the influential characteristics for every $X_\ell$ among those $X_k$ for which $\hat{\pi}(k) < \hat{\pi}(\ell)$. 
The idea is to detect those $f_{k, \ell}$ which are not $0$, using feature selection methods \cite{WoodBookGAM, huangSparseNonparametricRegression}. For those $k$, a change in $X_k$ has an effect on $X_\ell$. 
For a comparison of CAM and the PC algorithm based on 
simulated data sets with known ground truth, see \cite{CAM}.



\section{Methodology}\label{sec:Methodology}
The goal of this section is to derive a method that combines the current results on structure learning of SEMs with the features of the manufacturing domain in Section \ref{sec:Manufacturing}. 
\subsection{Recap of Common Prior Knowledge} \label{sec:commonSemanticInfo}
Compared to other applications of causal discovery, it is typical for the manufacturing domain, that there exists prior knowledge, see Section \ref{sec:Manufacturing}. In particular, there is a partial and transitive ordering of the variables implied by the stations' ordering. 
Additionally, we include expertise on the absence of edges. Both facets shall improve the algorithm's runtime. 
\subsection{Adaptions to CAM}\label{sec:TCAM}
The data generating process behind manufacturing data sets often leads to a low number of conditional independencies in $\mathbf{X}$, when compared to $p$. Thus, the Causal Bayesian Network of $\mathbf{X}$ is not sparse. This poses a challenge to many structural learning algorithms in higher dimensions. We show in this subsection how prior knowledge on the node ordering and the existence of edges can be incorporated so that structure learning remains feasible. To formalize our prior knowledge, let $t : \{1, \ldots, p\} \rightarrow \{1, \ldots, T\}$, so that $t(k) < t(\ell)$ means that there can only be edges from $X_k$ to $X_\ell$ but not vice versa. Further, let $F$ be a boolean matrix, where $F_{k, \ell} = \mathtt{True}$ if the edge from $X_k$ to $X_\ell$ is known to be absent. 
\subsubsection{Preliminary Neighborhood Selection}
For every measurement $X_\ell$, we determine a set of possible parents among those $k$, where $F_{k, \ell} = \mathtt{False}$ and $t(k) \leq t(\ell)$. Denote that set for index $\ell$ by $P_\ell$. 
\subsubsection{Node Ordering}
We start by adding all potential edges that go across stations and add them to the initial graph $G_0$, as those can not cause any cycle. The score of $G_0$ hence is 
\begin{equation}\label{eq:initScoreCAM}
S(G_0) = \sum_{\ell = 1}^p \sum_{k \in P_\ell, t(k) < t(\ell) } ||x_{\ell} - \widehat{f}_{k,\ell}(x_{k})||^2_{2, N}.
\end{equation}
We continue by determining the node ordering as in Section \ref{sec:graphLearningSEM}. Note that we only need to determine the node ordering for indices $k, \ell$ so that $t(k) = t(\ell)$. The initial inclusion of across-station-edges saves update steps of $M$ (Equation \Ref{eq:ScoreImprove}). This makes the algorithm feasible even for non-sparse high-dimensional settings, if the number of tiers $T$ or the number of edges known to be absent is sufficiently large. 
\subsubsection{Pruning}
The pruning step is identical to CAM. \\
In the manufacturing industry, the prior knowledge on $t(k)<t(l)$ is often given by the temporal nature of the production process. We therefore call our adaption \textit{TCAM} (\textit{Temporal} Causal Additive Models). It is sketched in Algorithm \ref{algo:TCAM}. 

\begin{algorithm}
\label{algo:TCAM}
\SetAlgoLined
\KwIn{$D$, $F$, $t$ as in Section \ref{sec:TCAM}}
\KwResult{DAG $G$} 
\tcp{Preliminary Neighborhood Selection (PNS)}
$\mathtt{SupersetNeighbors} = \mathtt{list()}$\;
\For{$\ell = 1, \ldots, p$}{
$I = \{k ~ s.t. ~ t(k) \leq t(\ell)  ~ \& ~ F(k,\ell) = \mathtt{False}\}$\;
$\mathtt{SupersetNeighbors}[\ell] = \mathtt{PNS}\left(X_\ell, \mathbf{X}_{I}\right)$\;
\tcp{Other edges are now forbidden}
\For{$k = 1, \ldots, p$}{
\If{$k \notin \mathtt{SupersetNeighbors}[\ell]$}{
$F(k, \ell) = \mathtt{True}$\;
}
}
}
\tcp{Add across-tier edges}
Set $G$ as empty graph on $X_1, \ldots, X_p$\;
\For{$k, \ell = 1, \ldots, p$}{
\If{$ k \in \mathtt{SupersetNeighbors}[\ell] ~ \& ~ t(k) < t(\ell)$}
{
Append $(k, \ell)$ to edges of $G$\;
}
}
$M(k, \ell) = $ right-hand side of (\ref{eq:initScoreCAM}) \;
\For{$k, \ell = 1, \ldots, p$}{
\If{$\left(\left(t(k) > t(\ell)\right) | \left(F(k, \ell) = \mathtt{True}\right)\right)$}{
$M(k, \ell) = -\infty $\;
}
}
\tcp{Add within-tier edges}
\While{$\max(M) > -\infty$}{
Find $(k_0, \ell_0) = \argmax_{(k,\ell) \in A(G)} M(k, \ell)$\;
Append $(k_0, \ell_0)$ to edges of $G$\;
$M(k_0, \ell_0) = -\infty$\;
Update $M(\cdot, \ell_0)$\;
Set $M(k, \ell) = -\infty$ for all $\{1, \ldots, p\} \times \{1, \ldots, p\} \ni (k, \ell) \notin A(G)$\; 
}
\tcp{Pruning like CAM (details omitted)}
\KwRet{$G$}
\caption{TCAM Algorithm}
\end{algorithm}

\section{Application}\label{sec:Application}
The energy storage of electric vehicles is called a battery pack which is composed of battery modules, which in turn contain a fixed number of battery cells. A battery module connects the battery cells in series or parallel and it protects those cells against shock, vibration and heat. Thus, the battery module is a key component for the safety of battery-electric vehicles. We apply TCAM to data collected at the assembly at BMW. The data set under investigation contains $7254$ battery modules with $738$ variables each. 
\subsection{Data Preparation}
As the missing values rate is low (around $2.4\%$) we apply naive mean imputation instead of more sophisticated method as \cite{MI, ramosaj2022relation, gaussCopula}. We then continue by removing features that have only one distinct value and hence provide no information. As this data set also shows multicollinearity, we apply an expert-based approach. We asked experts to identify clusters of variables containing similar information and to define representatives for them. For a purely data-driven approach in manufacturing, see \cite{marazopoulouCausal}. Those steps reduced the number of characteristics from $738$ to $491$. Finally, we standardize the data so the variables' empirical mean and standard deviation is $0$ and $1$ respectively. \\
Beyond the temporal ordering of the stations, it is reasonable that the production measurements of identical intermediate products as depicted in Figure \ref{fig:dataPrep} are independent. Thus, it is possible to restrict the potential edges that have to be considered. Additionally, we assume that some of the recorded measurements as the facility temperature and the selection of the stations are not affected by other measurements. We can mark those values as \textit{root nodes}, meaning that they have no incoming edges. This further restricts the number and orientation of possible edges. 
\subsection{Choice of Software and Hyperparameters}
\subsubsection{Preliminary Neighborhood Selection}
For our application of TCAM, we find supersets of the neighbors by applying the LASSO. 
For $\ell \in \{1, \ldots, p\}$, we run a regression of $X_\ell$ on those components of $\mathbf{X}$, which are possible parents according to our prior knowledge. Going forward we mark those variables as potential parents of $X_\ell$, where the corresponding regression coefficient is above $10^{-2}$. The penalty parameter $\lambda$ is chosen via cross-validation. Let $\lambda_{min}$ be the penalty parameter that minimizes the mean squared cross-validation error. Then we choose the maximal $\lambda$ such that the mean cross-validation error is within one standard deviation of the minimum $\lambda_{min}$. 
\subsubsection{Node Ordering and Pruning}
For the node ordering we employ the package \texttt{mgcv} by \cite{WoodBookGAM}. Let us call the graph after node ordering $G_{NO}$. In the pruning step we run a sparse additive regressions of $X_\ell$ on its parents in $G_{NO}$ for $\ell = 1, \ldots, p$. This step returns p-values for the parents of $X_\ell$ in $G_{NO}$. We follow \cite{CAM} and set the regressands as parents of $X_\ell$ in the final graph, whose p-values are below the threshold of $10^{-3}$.
\subsection{Results}
\begin{figure}
\includegraphics[width=\linewidth]{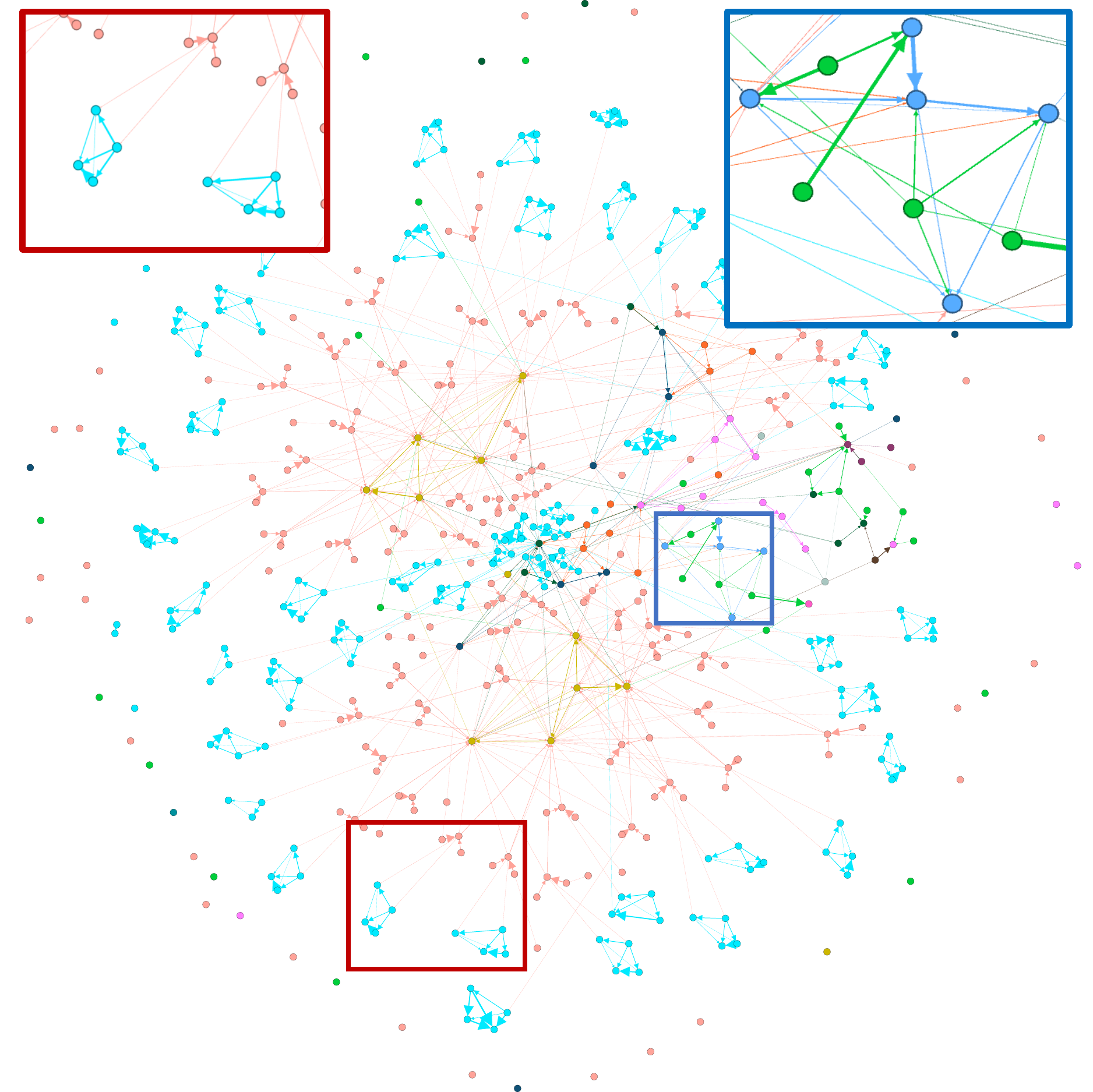}
\caption{Resulting graph of TCAM where the nodes correspond to characteristics of the product and edges correspond to detected CERs. The node coloring is according to the station, where the variable was measured. Edge colors are according to respective source node's color. The blue box highlights the detected relationship between the choice of the stations (green nodes) and the product quality (blue nodes). The red box depicts the similarities between structures of identical subcomponents.}
\label{fig:graph}
\end{figure}

The resulting graph is depicted in Figure \ref{fig:graph} and contains $491$ nodes and $859$ edges. We observe that there are a few nodes that have a large number of neighbors. In general this poses a difficulty for most structure learning algorithms and CAM did not finish in reasonable time. For details on the runtime for a low-dimensional special case, see Section \ref{sec:Subcomponent}. With TCAM and the inclusion of prior knowledge we were able to overcome those obstacles. \\
Further, substructures of identical parts show similar patterns. The red box in Figure \ref{fig:graph}, highlights patterns consisting of two linked clusters, where one cluster consists of four nodes, while the other one consists of three nodes. Together with process experts we could further verify that many CERs detected by TCAM are plausible. \\
This application is confidential, but we would still like to share one of the insights. TCAM discovered a CER between one station that processed the part and the part's quality. Experts derived that the maintenance of that station was overdue and the CER can be used to find better maintenance intervals. This is one example how graphical models can contribute to an effective and proactive process control.
\subsection{Evaluation against Expert Knowledge}\label{sec:Subcomponent}
\begin{figure}
\includegraphics[width=\linewidth]{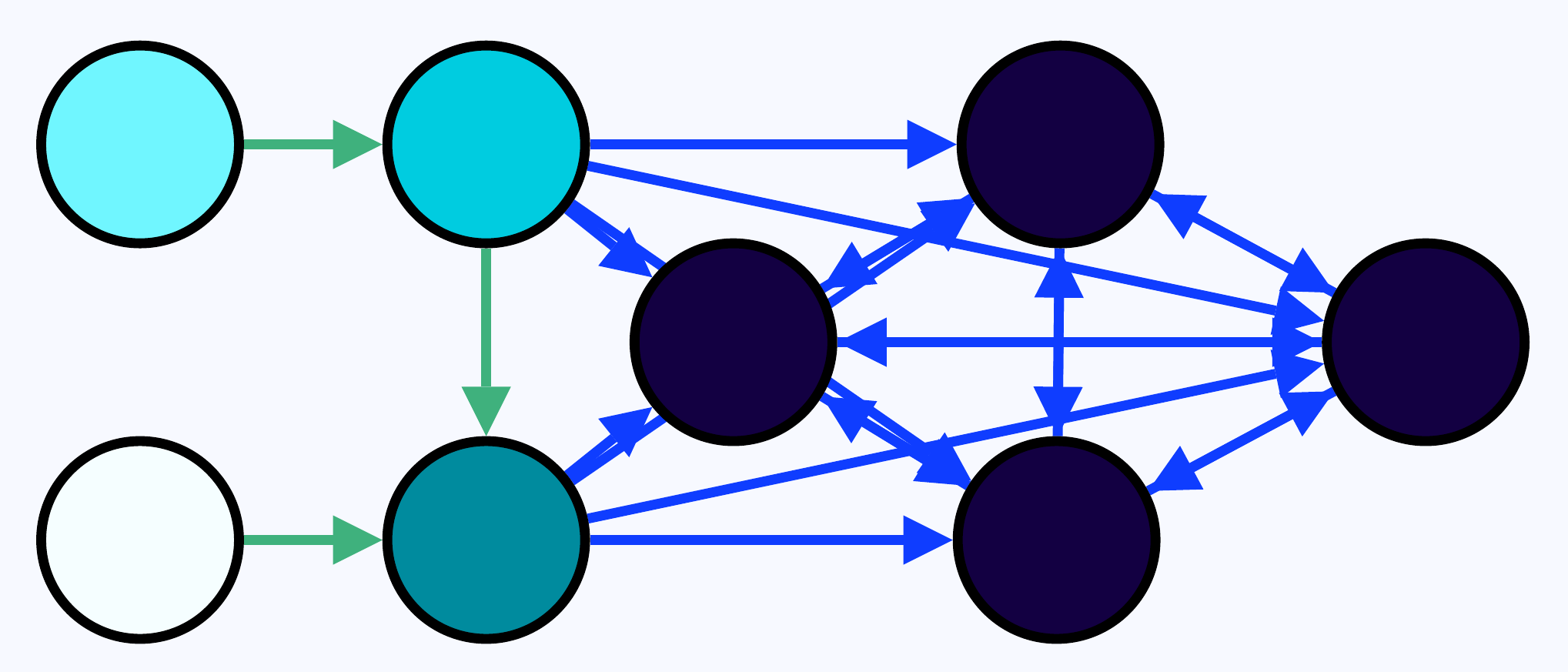}
\caption{Expert-based graph on measurements for subcomponents. The green edges are known to exist, while the blue edges potentially exist. Edges beyond the ones depicted are known to be absent. The darker the node, the later the corresponding variable is measured in the production process.}
\label{fig:graphCell}
\end{figure}

For the characteristics of one of the subcomponents, we derived an expert-based graph, which is depicted in Figure \ref{fig:graphCell}.
Here, the blue CERs potentially exist, while green CERs surely exist. Other CERs can be ruled out. We compare the estimated graphs and runtimes of TCAM, CAM and a variant of the PC algorithm called TPC \cite{TPC}, which allows the inclusion of temporal background knowledge. The significance level is set to $0.01$. We run $500$ experiments, where we randomly draw $500$ subcomponents, while each of them appears in at most one of the runs.  
\begin{table}
\begin{tabular}{c | c | c | c | c | c}
& $\overline{\text{aSHD}}$ & sd(aSHD) & $\overline{\text{\#edges}}$ & sd(\#edges) & $\overline{\text{time (s)}}$\\
\hline\hline
CAM & $3.496$  & $1.442$ & $9.464$ & $0.948$ & $1.342$\\
TCAM & $1.120$  & $0.343$ & $8.084$ & $0.778$ & $1.000$\\
TPC & $1.108$ & $0.866$ & $7.463$ & $1.623$ & $0.013$\\
\end{tabular}
\caption{The average aSHD ($\overline{\text{aSHD}}$), the standard deviation of the aSHD (sd(aSHD)), the average number of edges ($\overline{\text{\#edges}}$) and the standard deviation of the number of edges (sd(\#edges)) for all three methods of Section \ref{sec:Subcomponent} and for $500$ replications. 
}
\label{table:resultSubcomponent} 
\end{table}
We define an adapted Structural Hamming Distance (aSHD) \cite{shd} between an estimated graph $G_{est}$ and the one in Figure \ref{fig:graphCell} by the sum over the number of green edges that are not in $G_{est}$ and the number of edges $G_{est}$ that do not appear in Figure \ref{fig:graphCell}. The results are depicted in Table \ref{table:resultSubcomponent}. TPC and TCAM perform better than CAM, which shows the advantage of the inclusion of prior knowledge. Additionally, even in this low-dimensional setting the average runtime for TCAM is smaller than for CAM. Further, we observe that the aSHD of TCAM and TPC is on average quite similar. However, the standard deviation of the aSHD and the standard deviation of the number of edges is smaller for TCAM. This indicates that TCAM delivers more stable and informative results in the manufacturing domain. The original PC algorithm performed worse than TPC and is omitted.



\section{Conclusion}\label{sec:Conclusion}
We have presented a method to derive the graphical representation of CERs of manufacturing processes based on SEMs. While existing approaches for causal discovery in the manufacturing domain assumed linear relationships between the process characteristics, we applied CAM to find arbitrary additive functional relationships in data. We showed how existing prior domain knowledge can be included and improves the computational burden of CAM. A case study on manufacturing data reveals that the learned graph detects  unknown root-causes, delivers more informative results and paves the way to an efficient and proactive process control. \\


\bibliographystyle{IEEEtran}
\bibliography{IEEEabrv,bibliography}


\end{document}